\documentclass[a4paper,fleqn]{cas-dc}
\pdfoutput=1

\usepackage[numbers]{natbib}

\usepackage{joshuaeh}
\usepackage{amsmath,amssymb,amsfonts}
\usepackage{algorithmic}
\usepackage{algorithm}
\usepackage{graphicx}
\usepackage{textcomp}
\usepackage{xcolor}
\usepackage{booktabs}
\usepackage{multirow}
\usepackage{hyperref}
\usepackage{cleveref}
\usepackage{siunitx}
\usepackage{bm}
\usepackage{setspace}

\newcommand{\param}{\bm{\pi}}
\newcommand{\state}{\bm{x}}
\newcommand{\inp}{\bm{u}}
\newcommand{\cov}{\bm{P}}
\newcommand{\kalman}{\bm{K}}
\newcommand{\jacobian}{\bm{H}}
\newcommand{\src}{\mathcal{S}}
\newcommand{\tgt}{\mathcal{T}}
\newcommand{\scr}{^}  
\newcommand{\data}{\mathcal{D}}
\newcommand{\loss}{\mathcal{L}}
\newcommand{\pHorizon}{n_{out}}
\newcommand{\pInterval}{k + 1 \ldots k + \pHorizon}
\newcommand{\inInterval}{k \ldots k+\pHorizon-1}

\begin{document}
\let\WriteBookmarks\relax
\def\floatpagepagefraction{1}
\def\textpagefraction{.001}
\shorttitle{Using the SEKF to Transfer NN Models of Dynamical Systems with Limited Data}
\shortauthors{J.E. Hammond et~al.}

\title [mode = title]{Using the Subset Extended Kalman Filter to Adapt Pre-Trained Neural Network Models of Dynamical Systems with Limited Data}

\author[che]{Joshua E. Hammond}[orcid=0000-0001-7025-0150]
\ead{joshua.hammond@utexas.edu}
\credit{Conceptualization, Methodology, Software, Writing - Original Draft, Writing - Review \& Editing}

\author[che,ei,tmi]{Brian A. Korgel}[orcid=0000-0001-6242-7526]
\ead{korgel@che.utexas.edu}
\credit{Writing - Review \& Editing, Supervision, Funding acquisition}

\author[exxon]{Tyler A. Soderstrom}
\ead{tyler.a.soderstrom@exxonmobil.com}
\credit{Writing - Review \& Editing, Funding acquisition}

\author[che,oden]{Michael Baldea\corref{cor1}}[orcid=0000-0001-6400-0315]
\ead{mbaldea@che.utexas.edu}
\credit{Conceptualization, Methodology, Writing - Review \& Editing, Supervision, Funding acquisition}

\cortext[cor1]{Corresponding Author.}

\affiliation[che]{organization={McKetta Department of Chemical Engineering, The University of Texas at Austin},
    addressline={200 East Dean Keeton St., Stop C0400},
    city={Austin},
    postcode={78712},
    state={TX},
    country={United States}
}
\affiliation[ei]{organization={Energy Institute, The University of Texas at Austin},
    addressline={2304 Whitis Ave., Stop C2400},
    city={Austin},
    postcode={78712},
    state={TX},
    country={United States}
}
\affiliation[tmi]{organization={Texas Materials Institute, The University of Texas at Austin},
    addressline={204 E. Dean Keeton St., Stop C2201},
    city={Austin},
    postcode={78712},
    state={TX},
    country={United States}
}
\affiliation[exxon]{organization={ExxonMobil Technology and Engineering},
    city={Spring},
    state={TX},
    country={United States}
}
\affiliation[oden]{organization={Institute for Computational Engineering and Sciences, The University of Texas at Austin},
    addressline={201 E. 24th St., POB 4.102, Stop C0200},
    city={Austin},
    postcode={78712},
    state={TX},
    country={United States}
}

\begin{abstract}
Data-driven models of dynamical systems require extensive amounts of training data. For many practical applications, gathering sufficient data is not feasible due to cost or safety concerns. This work uses the Subset Extended Kalman Filter (SEKF) to adapt pre-trained neural network models to new, similar systems with limited data available. Experimental validation across damped spring and continuous stirred-tank reactor systems demonstrates that small parameter perturbations to the initial model capture target system dynamics while requiring as little as 1\% of original training data. In addition, finetuning requires less computational cost and reduces generalization error.
\end{abstract}



\maketitle

\section{Introduction}
\label{sec:intro}

Data-driven models have emerged as powerful tools for approximating the behavior of complex dynamical systems across diverse domains. Artificial Neural Networks (ANNs), in particular, demonstrate remarkable capability in learning nonlinear mappings from arbitrary inputs to outputs. For example, ANN models can effectively capture dynamics governed by ordinary differential equations of the form $\frac{d\state}{dt} = f(\state, \inp, p)$ without requiring explicit knowledge of the underlying physics~\cite{chen2018neural}. However, these data-driven approaches suffer from fundamental limitations that impede their widespread deployment in practical applications, particularly industrial settings. Neural network models provide no guarantee of performance outside the range of the training data, exhibit poor generalization when the statistical properties of operating conditions change from those encountered during training, and require extensive datasets to achieve satisfactory generalization. The acquisition of sufficient training data presents substantial barriers in many practical applications, where data collection may be constrained by safety considerations, operational costs, or time limitations.

Transfer learning offers a principled approach to address data scarcity by leveraging knowledge acquired from data-rich source domains to improve learning in data-scarce target domains. This paradigm has achieved remarkable success in computer vision and natural language processing, where established practices include layer-wise freezing and fine-tuning of pre-trained networks. The conventional methodology exploits hierarchical feature representations: early layers encode generic, transferable features while later layers capture task-specific patterns, providing natural guidance for selective parameter adaptation. However, the direct application of such techniques to neural network models of dynamical systems presents challenges that remain largely unaddressed. Unlike image classification, where discrete architectural layers correspond to progressively abstract feature hierarchies, dynamical systems models lack clear analogues for layer-wise feature extraction, complicating the identification of which parameters to adapt during transfer learning. Furthermore, existing approaches predominantly rely on gradient-based optimization methods that lack probabilistic frameworks to mitigate overfitting when target data is severely limited.

Four fundamental questions thus arise. First, whether models trained on functionally similar source dynamical systems can be effectively leveraged to improve target system performance when data from the target system is severely limited remains unclear. Second, the extent to which small parameter perturbations to a source system model can capture target system dynamics---particularly when systems differ through parametric variations in governing equations---has not been systematically characterized. Third, principled strategies for parameter adaptation that mitigate overfitting to scarce target observations are notably absent from the existing literature. This knowledge gap is particularly striking given the prevalence of parametric variability in industrial applications, including unit-to-unit differences in thermal systems, and simulation-to-real transfer scenarios. Fourth, whether there exist heuristics or principles to guide which ANN model parameters should be modified during transfer learning for dynamical systems remains an open question. Unlike computer vision, where layer-wise feature hierarchies inform parameter selection, no analogous framework exists for dynamical systems models, complicating the design of effective transfer learning strategies.

This work addresses these questions through a transfer learning framework based on the Subset Extended Kalman Filter (SEKF) for neural network models of dynamical systems~\cite{hammond2025selective}. The key hypothesis is that transfer learning can be formulated as Bayesian inference: when the two systems are similar, the source model parameters $\param\scr\src$ define a Gaussian prior distribution $p(\param) = \mathcal{N}(\param\scr\src, \cov_0)$ over target model parameters. Adaptation then proceeds through sequential Bayesian updating as target observations become available. The SEKF provides an efficient mechanism for computing the posterior $p(\param\scr\tgt | \data\scr\tgt, \param\scr\src)$, with the process noise covariance $\bm{Q}$ controlling prior flexibility and the measurement noise covariance $\bm{R}$ weighting observation reliability. Unlike gradient-based optimization methods that lack explicit prior information, the SEKF maintains estimates of parameter uncertainty through covariance propagation, enabling principled regularization in data-limited regimes.

Experimental validation across two benchmark systems---a damped spring system and a Temperature Control Lab (TCLab) experiment---yields four principal findings. First, small parameter changes to source models suffice to capture target system dynamics: cosine similarities between source and adapted parameters exceed an average value of 99\% across all experimental conditions. Second, fine-tuning source model parameters via SEKF to reflect the target system substantially outperforms retraining from random initialization when target data availability is limited. Fine-tuning the target model achieves accuracy comparable to the source system model with as little as 1\% of the original training dataset. Third, the SEKF approach reduces overfitting relative to gradient-based optimization, as evidenced by smaller train-test performance gaps. Fourth, contrary to established practices in computer vision, parameter changes distribute across all network layers rather than concentrating in output layers, suggesting that transfer learning for dynamical systems requires coordinated adaptation throughout the architecture. These results establish the SEKF framework as a principled approach for developing system-specific data-driven models when training observations are severely constrained.

\section{Related Works}\label{sec:related_works}

Transfer learning has achieved remarkable success in domains such as computer vision and natural language processing, where pre-trained models are adapted to new tasks with limited data. A key challenge in these applications is mitigating overfitting to the small target datasets, which is addressed by modifying only a subset of model parameters during fine-tuning. For instance, it is common practice to freeze the weights of earlier layers and retrain only the final layers when performing transfer learning on deep computer vision ANN classification models. This approach is guided by the intuition that earlier layers capture generalizable features, such as edges and textures, that are transferable across tasks, while the later layers learn task-specific combinations of these features corresponding to different classes. This parameter-efficient strategy has enabled the widespread adoption of transfer learning in scenarios where data is scarce.

\subsection{Transfer Learning for Dynamical Systems}

While transfer learning has achieved remarkable success in computer vision and natural language processing, its application to dynamical systems and regression remains relatively underexplored. Forgione et al.~\cite{forgione2023system} introduced Jacobian Feature Regression for rapid recurrent neural network adaptation when system dynamics change, demonstrating parameter-efficient adaptation through feature-space modifications. Alhajeri et al.~\cite{alhajeri2024transfer} extended this paradigm to chemical process control, applying transfer learning-based recurrent neural networks to model predictive control with limited operational data. However, both approaches rely on gradient-based fine-tuning methods applied to all ANN parameters that lack probabilistic frameworks to mitigate overfitting, representing a critical limitation when target system data is severely constrained.

Unlike transfer learning for computer vision, there is no heuristic or intuition to guide which ANN model parameters should be modified. The question of which network layers to adapt during transfer learning has received attention in the context of fluid dynamics. Chatzi et al.~\cite{chatzi2023spectral} provided spectral analysis frameworks for understanding layer-wise retraining in turbulence modeling, revealing that optimal adaptation strategies differ substantially between dynamical systems and traditional image classification tasks. Nevertheless, systematic investigations of how parameter changes distribute across network architectures in regression-based dynamical system transfer learning remain absent from the literature.

\subsection{Neural Network System Identification}

Foundational work on neural ordinary differential equations by Chen et al.~\cite{chen2018neural} established continuous-depth architectures as natural representations for dynamical systems, enabling direct modeling of continuous time dynamics $\frac{d\state}{dt} = f(\state, \inp, \theta)$. Kutz and Brunton~\cite{kutz2022machine} provide a comprehensive review of machine learning methods for dynamical system prediction, highlighting persistent challenges in generalization beyond training distributions and data efficiency.

Physics-informed approaches have emerged to incorporate domain knowledge when data is scarce. Chatzi and Naseri~\cite{chatzi2021discrepancy} proposed discrepancy modeling frameworks that decompose neural ordinary differential equations into physics-informed and purely data-driven components. Velioglu et al.~\cite{velioglu2024physics} demonstrated physics-informed neural networks for continuously stirred tank reactor modeling with limited data, building on the foundational framework of Raissi et al.~\cite{raissi2019physics}. Wang and Wu~\cite{wang2024meta} combined meta-learning with physics-informed fine-tuning to enable rapid adaptation for chemical reactor modeling. While these methods reduce data requirements through physics constraints, they require explicit knowledge of governing equations, an assumption that does not hold for many scenarios where model form uncertainty dominates.

\subsection{Kalman Filtering for Neural Network Training}

The Extended Kalman Filter offers a probabilistic alternative to gradient-based optimization, treating neural network parameters as states to be estimated. Haykin's~\cite{haykin2001kalman} seminal work established theoretical foundations for EKF-based neural network training, framing parameter updates as Bayesian posterior estimation. Recent advances include KalmanNet~\cite{revach2022kalmannet}, which hybridizes model-based Kalman filtering with data-driven recurrent neural networks for systems with partially known dynamics, and differentiable Kalman filter frameworks~\cite{coskun2017long} that enable end-to-end learning. Chang~\cite{chang2023low} introduced low-rank EKF approximations for online neural network learning from streaming data, addressing computational tractability for large-scale models. However, these methods focus primarily on online learning scenarios rather than transfer learning contexts, and none systematically investigate how Kalman filtering-based approaches compare to gradient methods in mitigating overfitting when adapting pre-trained models with minimal target data.

\subsection{Parameter-Efficient Fine-Tuning and Overfitting Reduction}

In natural language processing and computer vision, parameter-efficient fine-tuning methods have demonstrated that small, targeted parameter modifications suffice for effective domain adaptation. LoRA~\cite{hu2022lora} showed that low-rank updates to weight matrices enable adaptation of large language models without full retraining. Ding et al.~\cite{ding2023peft} established that parameter-efficient fine-tuning inherently reduces overfitting through implicit regularization mechanisms. In scientific computing domains, K\"aser et al.~\cite{kaser2019layer} found that when predicting chemical properties, finetuning the later layers of the ANN model yields better generalization than full finetuning when the amount of data from the target domain is limited.

From a Bayesian perspective, parameter-efficient methods implicitly place strong priors on frozen parameters (effectively infinite precision, constraining them exactly to source values) and weak priors on adapted parameters. Layer-freezing heuristics in computer vision thus encode prior beliefs about which parameters are transferable. However, for dynamical systems where no clear layer-wise hierarchy exists, principled methods for encoding prior uncertainty across all parameters---rather than binary freeze / unfreeze decisions---remain unexplored. Whether Bayesian optimization frameworks that maintain full covariance structures offer advantages over gradient-based fine-tuning in this context has not been systematically investigated.

\subsection{Gaps Addressed by This Work}

Despite these advances, four critical gaps persist. First, whether models trained on functionally similar source dynamical systems can be effectively leveraged to improve target system performance when data from the target system is severely limited remains unclear. Second, the extent to which small parameter perturbations to a source system model can capture target system dynamics---particularly when systems differ through parametric variations in governing equations---has not been systematically characterized. Third, principled strategies for parameter adaptation that mitigate overfitting to scarce target observations are notably absent from the existing literature.  Fourth, whether there exist heuristics or principles to guide which ANN model parameters should be modified during transfer learning for dynamical systems remains an open question

This addresses these gaps using the Subset Extended Kalman Filter (SEKF) to transfer ANN models of source dynamical systems to target systems. We demonstrate that small parameter perturbations embedded within a probabilistic updating scheme enable effective transfer learning for dynamical systems with as little as 1\% of original training data, while inherently mitigating overfitting through Bayesian noise modeling.

\section{Transfer Learning of Dynamical Systems}
\label{sec:transfer_learning_dynamical_systems}

\subsection{System Definition}

Consider a source dynamical system governed by ordinary differential equations of the form

\begin{equation}
    \frac{d\state}{dt} = f(\state, \inp, p),
    \label{eq:source_system}
\end{equation}

where $\state \in \R^{n_x}$ denotes the state vector, $\inp \in \R^{n_u}$ represents the input or manipulated variable vector, and $p \in \R^{n_p}$ comprises the physical parameters of the source system. The dynamics described by Eq.~\eqref{eq:source_system} are assumed to be well-posed, admitting unique solutions for specified initial conditions $\state_0$ and input trajectories within the operating domain of interest.

A neural network surrogate model $\phi: \R^{n_x} \times \R^{n_u} \times \R^{n_\pi} \to \R^{n_x \times n_{out}}$ approximates the discrete-time evolution of the source system at time $k$ across $\pHorizon$ predictions according to

\begin{equation}
    \tilde\state_{\pInterval} = \phi(\state_k, \inp_{\inInterval}, \param),
    \label{eq:nn_surrogate}
\end{equation}

The neural network $\phi$ may comprise various architectures such as multi-layer perceptrons (MLPs), recurrent neural networks (RNNs), or neural ordinary differential equations (NODEs), with the specific choice depending on the complexity of the underlying dynamics and computational requirements. The parameters of the neural network $\param \in \R^{n_\pi}$ are optimized using data from the source system $\data\scr\src = \{(\hat\state_k, \hat\inp_k, \hat\state_{k+1\ldots k+n_{out}})\}_{k=1}^{N\scr\src}$. Training the neural network weights minimizes a loss function $\loss$ such as mean squared error that quantifies the discrepancy between predicted $\tilde\state_{\pInterval}$ and measured states $\hat\state_{\pInterval}$.

Transfer learning aims to model a target $\tgt$ dynamical system that is similar to the source $\src$ system but has significantly less training data available $N\scr\tgt \ll N\scr\src$. If the two systems are sufficiently similar, it is hypothesized that there exists a set of optimal neural network parameters $\param\scr\tgt$ that can be obtained by adapting the previously-trained source parameters $\param\scr\src$ using only the limited target data $\data\scr\tgt$. Ideally, this process of \emph{finetuning} yields a target model that performs better than re-initializing and \emph{retraining} the neural network from scratch using only $\data\scr\tgt$. Table~\ref{tab:system_definitions} summarizes the definitions of the source and target systems used in this work and how their respective spaces relate to one another.

\begin{table*}
    \centering
    \caption{Summary of source and target dynamical systems used for transfer learning experiments.}
    \label{tab:system_definitions}
    \begin{tabular}{llll}
        \toprule
        {} & \textbf{Source System} & \textbf{Target System} & {}\\
        \midrule
        {System Space} & $\frac{d\state}{dt} = f\scr\src(\state, \inp, p\scr\src)$ & $\frac{d\state}{dt} = f\scr\tgt(\state, \inp, p\scr\tgt)$ & {}\\
        {Data Space} & $\data\scr\src$ & $\data\scr\tgt$ & $\dim\data\scr\src \gg \dim\data\scr\tgt$ \\
        Model Space & $\tilde\state_{\pInterval} = \phi\scr\src(\state_k, \inp_{\inInterval}, \param\scr\src)$ & $\tilde\state_{\pInterval} = \phi\scr\tgt(\state_k, \inp_{\inInterval}, \param\scr\tgt)$ & $\phi\scr\src$ and $\phi\scr\tgt$ have identical structures \\
    \end{tabular}
\end{table*}

\begin{example}
    \label{ex:damped_spring_mass}
    Consider a damped spring mass system governed by the second-order ordinary differential equation $m\ddot{x} + c\dot{x} + kx = u$, where $m$ is the mass, $c$ is the damping coefficient, $k$ is the spring constant, and $u$ is a constant external forcing function. A neural network surrogate model $\phi$ is trained to predict the position at each second over the next $\pHorizon$ 20 seconds $x_{1\ldots20}$ given the initial position and velocity $[x_0, \dot{x}_0]$. Figure~\ref{fig:example_trajectory} illustrates an example trajectory of the system given the initial conditions, with the blue $+$ markers denoting measurements and model predictions shown with the orange $\times$ markers.

    \begin{figure}
        \centering
        \includegraphics[width=\columnwidth]{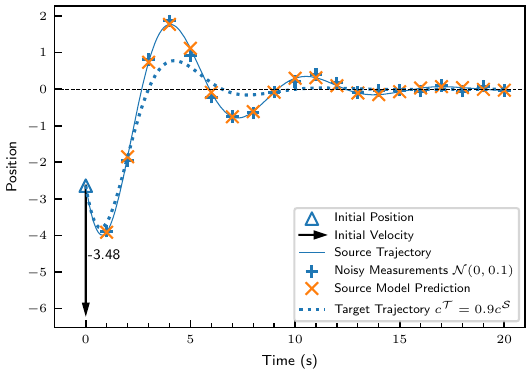}
        \caption{Example trajectory of a damped spring-mass system (solid line) given initial position (blue triangle) and velocity (black arrow). A neural network trained on noisy measurements (blue $+$) makes predictions (orange $\times$) of future positions given the initial position and velocity. A target system with slightly different damping coefficient (dashed line) illustrates the transfer learning objective.}
        \label{fig:example_trajectory}
    \end{figure}

    Now consider a target system that is identical to the source system except that the damping coefficient is reduced by 10\%, i.e., $c^\tgt = 0.9 c^\src$. Figure~\ref{fig:example_trajectory} shows this trajectory with a dashed line. The objective of transfer learning is to adapt the neural network parameters $\param\scr\src$ trained on the source system to accurately predict the dynamics of the target system using only a limited dataset $\data\scr\tgt$ collected from the target system. Successful transfer learning would enable accurate prediction of the target system's trajectories with minimal additional data and computational effort.

\end{example}

\subsection{Assumptions}

The transfer learning framework developed in this work relies on the following assumptions:

\begin{enumerate}
    \item \textbf{Functional Similarity}: The source and target systems share the same functional form of governing equations. This assumption ensures that the representational capacity required to model both systems is similar. In contexts where the governing equations are unknown, comparing the trajectories generated by the two systems given the same input can provide empirical evidence of functional similarity.

    \item \textbf{Data Domain Overlap}: The state and input variables of the target system occupy a similar range and scale as those of the source system, i.e., the operating domains substantially overlap. This assumption ensures that the features learned by the source model remain relevant for the target system.

    \item \textbf{Sufficient Source Training}: The source model has been trained to convergence on a sufficiently large and representative dataset such that $\param_s$ constitutes a high-quality approximation of the source system dynamics.

    \item \textbf{Limited Target Data}: The target dataset $\mathcal{D}_t$ is insufficient for training a neural network from random initialization to satisfactory performance, motivating the transfer learning approach.
\end{enumerate}

These assumptions define the scope of applicability for the proposed methodology. Systems violating these assumptions, such as those with substantially different governing equations or non-overlapping operating domains, may require alternative approaches beyond the framework presented here.

\subsubsection{Transfer Learning via Gradient Descent}
\label{subsec:gradient_descent}

Gradient-based fine-tuning adapts the pre-trained source parameters $\param\scr\src$ to the target system by minimizing the mean squared prediction error on the target dataset:

\begin{equation}
    \text{MSE} (\param) = \frac{1}{N} \sum_{i=1}^{N} \left\| \hat{\state}_{i} - \phi(\hat{\state}_{i-1}, \hat{\inp}_{i-1}, \param) \right\|_2^2.
    \label{eq:mse_loss}
\end{equation}

\begin{equation}
    \param\scr\tgt = \argmin_{\param} \text{MSE} (\param)
    \label{eq:optimization_problem}
\end{equation}
Three optimization algorithms are employed in this work. We modify the Adam optimizer~\cite{kingma2014adam} to update a subset of model parameters that have the highest gradient with respect to the loss function. More details on this method are available in \cite{hammond2025selective}. The Limited-memory Broyden-Fletcher-Goldfarb-Shanno (L-BFGS)\cite{byrd1995limited} algorithm approximates curvature information using gradient history, offering accelerated convergence for smooth objectives at the cost of increased memory requirements.

For transfer learning, parameters are initialized as $\param_0 = \param\scr\src$, and optimization proceeds via mini-batch stochastic gradient descent with early stopping based on validation loss. For comparison, retraining experiments initialize parameters from the same random distribution used for source model training. Hyperparameters including learning rate, batch size, and learning rate scheduling are determined through the ASHA algorithm~\cite{li2020system} as detailed in Appendix~\ref{app:exp_details}.

\subsection{Transfer Learning via Subset Extended Kalman Filter}
\label{subsec:sekf}

Transfer learning can be cast as Bayesian inference over neural network parameters. Let $p(\param\scr\tgt | \data\scr\src)$ denote the prior distribution over target parameters informed by source training. Under the assumption of functional similarity, we model this prior as Gaussian centered at the source parameters: $p(\param\scr\tgt) = \mathcal{N}(\param\scr\src, \cov_0)$, where $\cov_0$ encodes uncertainty about parameter transferability. Given target observations $\data\scr\tgt$, the goal is to compute the posterior $p(\param\scr\tgt | \data\scr\tgt, \param\scr\src)$. The Extended Kalman Filter (EKF)~\cite{hammond2025selective} provides an efficient sequential approximation to this posterior through recursive application of Bayes' rule, treating neural network parameters as hidden states to be estimated from noisy observations~\cite{haykin2001kalman}. This formulation casts parameter adaptation as a state estimation problem:
\begin{subequations}
\label{eq:ekf_state_space}
\begin{align}
    \param_{k+1} &= \param_k + \bm{w}_k \label{eq:process_model} \\
    \bm{w}_k &\sim \mathcal{N}(\bm{0}, \bm{Q})\\
    \tilde{\state}_{\pInterval} &= \phi(\hat{\state}_k, \hat{\inp}_{\inInterval}, \param_k) + \bm{v}_k, \label{eq:measurement_model} \\
    \bm{v}_k &\sim \mathcal{N}(\bm{0}, \bm{R})
\end{align}
\end{subequations}
where Eq.~\eqref{eq:process_model} models parameter evolution with process noise covariance $\bm{Q} \in \R^{n_\pi \times n_\pi}$, and Eq.~\eqref{eq:measurement_model} relates parameters to observed state transitions with measurement noise covariance $\bm{R} \in \R^{n_x \times n_x}$. The EKF recursively updates parameter estimates through prediction and correction steps involving the Jacobian $\jacobian_k = \partial \phi / \partial \param$ evaluated at the current estimate.

This formulation admits a Bayesian interpretation. Equation~\eqref{eq:process_model} models epistemic uncertainty: parameters may evolve from their source values, with $\bm{Q}$ quantifying expected deviation---larger values indicate weaker prior constraints. Equation~\eqref{eq:measurement_model} models aleatoric uncertainty: observations are corrupted by noise with covariance $\bm{R}$. The Kalman gain $\kalman_k = \cov_k \jacobian_k^\top (\jacobian_k \cov_k \jacobian_k^\top + \bm{R})^{-1}$ optimally balances prior information (encoded in the covariance $\cov_k$) against new observations, automatically providing stronger regularization when target data is scarce or noisy.

Direct application of the EKF to neural networks is computationally prohibitive due to the $\mathcal{O}(n_\pi^2)$ storage and $\mathcal{O}(n_\pi^3)$ inversion costs associated with the covariance matrix. The Subset Extended Kalman Filter (SEKF) addresses this limitation by updating only a subset $\mathcal{I}_k \subset \{1, \ldots, n_\pi\}$ of parameters at each step, with $|\mathcal{I}_k| = m \ll n_\pi$~\cite{haykin2001kalman}. The algorithm maintains and updates only the $m \times m$ submatrix of the covariance corresponding to selected parameters.

For transfer learning, the SEKF initializes $\param_0 = \param\scr\src$ with initial covariance $\cov_0 = \sigma_0^2 \bm{I}$, encoding the source parameters as a Bayesian prior. The process noise $\bm{Q}$ controls regularization strength: small values constrain parameters near the source model, while larger values permit greater adaptation. This probabilistic framework provides implicit regularization against overfitting by weighting new observations against prior information encoded in the covariance structure. The parameters $\sigma_0^2$ and $\bm{Q}$ are treated as hyperparameters and are tuned using methods described in Appendix~\ref{app:exp_details}. More detail on the SEKF is available in \cite{hammond2025selective}.

\subsection{Evaluation Metrics}
\label{subsec:metrics}

Evaluating the effectiveness of transfer learning requires metrics that quantify both model accuracy and generalization. When training, we use Mean Squared Error (MSE) loss as defined in Eq.~\eqref{eq:mse_loss}. Loss $\loss$ is computed for the training data $\data$ as well as for validation and test datasets. The validation error is used during training and hyperparameter tuning to prevent overfitting while the test error is used for final model assessment and approximation of generalization error. Since mitigating overfitting to the limited target data, and improving generalization is the primary goal (and challenge) in transfer learning, specifically note the \emph{Train-Test Gap} as the difference between the loss on the training data and the loss on the held-out test data as shown in Eq.~\eqref{eq:train_test_gap}. More details on model training are available in Appendix~\ref{app:exp_details}.

\begin{equation}
    \text{Train-Test Gap} = \loss(\data_{Test}^\tgt) - \loss(\data_{Train}^\tgt).
    \label{eq:train_test_gap}
\end{equation}

We use values from the source system to normalize these metrics for better interpretability across different systems and scales. Specifically, we report the \emph{Normalized MSE} and \emph{Normalized Convergence time} as shown in Eqs.~\eqref{eq:normalized_mse} and~\eqref{eq:normalized_convergence_time} respectively.

\begin{equation}
    \text{Normalized MSE} = \frac{\loss(\data_{Test}^\tgt)}{\loss(\data_{Test}^\src)}.
    \label{eq:normalized_mse}
\end{equation}

\begin{equation}
    \text{Normalized Convergence Time} = \frac{t^\tgt_{converge}}{t^\src_{converge}}.
    \label{eq:normalized_convergence_time}
\end{equation}

Finally, our assumptions dictate that the source and target systems are functionally similar, and that the final models should therefore have similar parameters. We quantify the similarity of the source and target \emph{data distributions} using the \emph{Wasserstein Distance} as shown in Eq.~\eqref{eq:wasserstein_distance}. There, $P$ denotes the empirical distribution of the data $\data$ and $\Gamma$ denotes the set of all joint distributions with marginals $P(\data^\src)$ and $P(\data^\tgt)$. We compare the \emph{functional similarity} of the source and target systems using Cosine Similarity (CS) as shown in Eq.~\eqref{eq:cosine_similarity} between trajectories generated by both systems given the same input sequences. Similarly, we use CS to compare the similarity of source and target model parameters after transfer learning. Table \ref{tab:similarity_metrics} summarizes all metrics used to evaluate transfer learning performance.

\begin{equation}
    W(P(\data^\src), P(\data^\tgt)) = \inf_{\gamma \in \Gamma(P(\data^\src), P(\data^\tgt))} \mathbb{E}_{(x,y) \sim \gamma} [\| x - y \|]
    \label{eq:wasserstein_distance}
\end{equation}

\begin{equation}
    \text{cos}(\param\scr\src, \param\scr\tgt) = \frac{\param\scr\src \cdot \param\scr\tgt}{\|\param\scr\src\|_2 \|\param\scr\tgt\|_2}.
    \label{eq:cosine_similarity}
\end{equation}

\begin{table*}
\centering
\caption{Metrics to quantify similarity between source and target systems.}
\label{tab:similarity_metrics}
\begin{tabular}{ll}
\toprule
\textbf{Metric} & \textbf{Purpose} \\
\midrule
Mean Squared Error (MSE) & Quantify model prediction error \\
Train-Test Gap & Measure overfitting and generalization \\
Normalized MSE & Compare target model accuracy to source model \\
Normalized Convergence Time & Compare training efficiency to source model \\
Wasserstein Distance & Quantify similarity of source and target data distributions \\
Cosine Similarity (CS) & Quantify similarity of source and target functional spaces and models \\
\bottomrule
\end{tabular}

\end{table*}

\section{Experiments}
\label{sec:experiments}

We evaluate how transfer learning (i.e. finetuning source model parameters) compares to retraining from random initialization on two benchmark dynamical systems: (1) the damped spring system described in Example~\ref{ex:damped_spring_mass} and (2) a Temperature Control Laboratory (TCLab) system~\cite{park2020benchmark,de2020introducing} a small experimental system consisting of two heaters and two temperature sensors in a small enclosure. Using a year of simulated data, a NN model is trained to predict heater temperatures across a ten-minute prediction horizon given initial temperatures, and heater settings across the prediction horizon. While the damped spring system provides a simple, well-understood test case for validating the transfer learning framework, the TCLab case demonstrates the nearly-ubiquitous scenario of transferring models from simulated to physical systems where data is limited. A notable practical advantage of the finetuning approach---particularly when using the SEKF---is that target data can be gathered \emph{online} during normal system operation rather than requiring dedicated, discrete experiments. Because the SEKF processes observations sequentially, the model can be adapted continuously as new operational data becomes available, avoiding the need for disruptive model redeployment cycles that batch gradient-based methods would require.

In addition to comparing parameter initialization methods, we compare the effect of the optimizer choice and amount of data available from the target system. A summary of the experimental design is provided in Table~\ref{tab:experiment_summary} with more details available in Appendix~\ref{app:supp_data}.

\begin{table*}
\centering
\caption{Summary of TL experiments}
\label{tab:experiment_summary}
\begin{tabular}{lll}
\toprule
Manipulated\\
\midrule
Parameter Initialization: & Finetune, Retrain \\
Optimization Methods: & SEKF, Adam, LBFGS \\
Dataset Sizes: & 10, 50, 100, 500, 1,000 (Damped Spring)\\
{} & 0.5 hr, 1 hr, 4 hr, 12 hr, 24 hr (10-second sampling frequency) (TCLab) \\
Transfer: & $m$,$k$,$c$,$u$ $\pm 10\%$ (Damped Spring) \\
{} & Simulated $\rightarrow$ Real (TCLab) \\
Replicates: & 10 (Damped Spring) \\
{} & 5 (TCLab) \\
\bottomrule
\end{tabular}
\end{table*}

\subsection{Results}

\begin{figure}[htbp]
\centering
\includegraphics[width=\columnwidth]{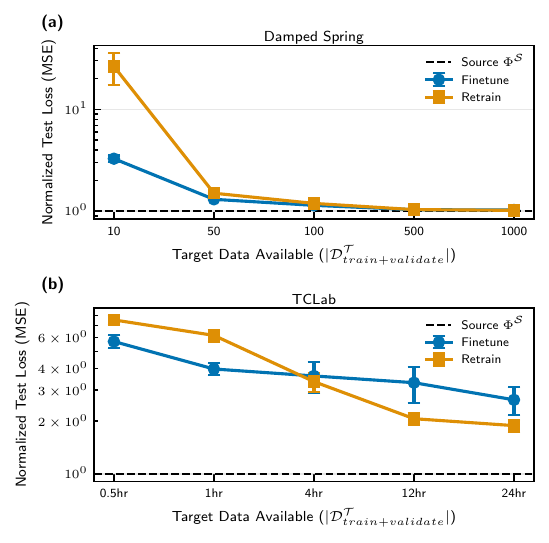}
\caption{Normalized test loss versus target data size for finetuning and retraining initialization methods. \textbf{(a)} Damped spring system. \textbf{(b)} TCLab system. Error bars indicate standard error across replicates. The dashed line indicates source model performance. Finetuning achieves lower test loss, especially with limited data.}
\label{fig:test_loss_comparison}
\end{figure}

Figure~\ref{fig:test_loss_comparison} shows normalized test loss as a function of target data availability for both initialization methods across both benchmark systems. Finetuning consistently achieves lower test loss than retraining, with the advantage being most pronounced at small dataset sizes. As more target data becomes available, the performance gap narrows, confirming that transfer learning provides the greatest benefit in data-scarce scenarios. For the damped spring system, the GLM interaction analysis (Section~\ref{sec:interaction}) shows that finetuning and retraining reach effectively equivalent performance at approximately 1,000 samples. For the TCLab system, the convergence occurs sooner---the interaction terms are not individually significant, suggesting that finetuning's advantage diminishes more rapidly with increasing data in the noisier, real-world setting.

\begin{figure}[htbp]
\centering
\includegraphics[width=\columnwidth]{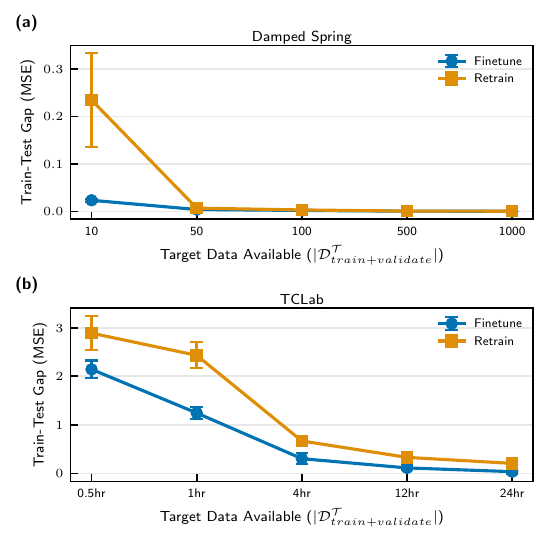}
\caption{Train-test gap (test loss minus training loss) versus target data size, grouped by initialization method. Smaller gaps indicate better generalization. Finetuning exhibits consistently smaller train-test gaps than retraining across all data sizes in both systems.\textbf{(a)} Damped spring system. \textbf{(b)} TCLab system.}
\label{fig:train_test_gap}
\end{figure}

The regularization benefit of finetuning is evident in the train-test gap comparison shown in Figure~\ref{fig:train_test_gap}. Retraining from random initialization produces larger gaps between training and test performance, indicating greater overfitting. Finetuning constrains optimization to remain near the well-generalized source parameters, resulting in models that generalize better to unseen data.

\begin{figure}[htbp]
\centering
\includegraphics[width=\columnwidth]{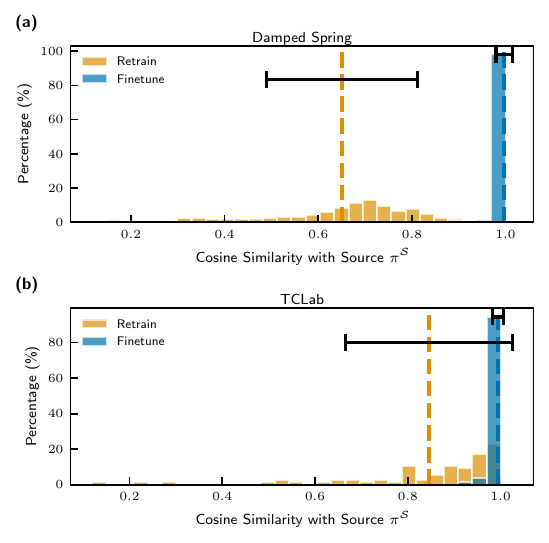}
\caption{Distribution of cosine similarity between adapted model parameters and source model parameters. Dashed vertical lines indicate mean cosine similarity for each initialization method, with horizontal error bars showing $\pm 1$ standard deviation. \textbf{(a)} Damped spring system. \textbf{(b)} TCLab system. Finetuned models cluster tightly near 1.0, indicating that successful adaptation requires only small perturbations from the source model. Retrained models exhibit much wider spread and lower similarity, occupying different regions of parameter space.}
\label{fig:cosine_similarity}
\end{figure}

Figure~\ref{fig:cosine_similarity} demonstrates that finetuned model parameters maintain high cosine similarity with source parameters, typically exceeding 0.99. This confirms that effective transfer learning for dynamical systems occurs within a small neighborhood of the source parameter space, validating our assumption of functional similarity between source and target systems.

\begin{table}[htbp]
\centering
\caption{Average cosine similarity between source and target parameters, broken down by system, initialization method, and optimizer.}
\label{tab:cosine_similarity}
\begin{tabular}{llcc}
\toprule
\textbf{System} & \textbf{Method} & \textbf{Optimizer} & \textbf{Cosine Similarity} \\
\midrule
Damped Spring & Finetune  & Adam    & $0.9959 \pm 0.0113$ \\
              &           & L-BFGS  & $0.9984 \pm 0.0156$ \\
              &           & SEKF    & $0.9977 \pm 0.0238$ \\
              & Retrain   & Adam    & $0.6519 \pm 0.1205$ \\
              &           & L-BFGS  & $0.6819 \pm 0.1210$ \\
              &           & SEKF    & $0.6047 \pm 0.2385$ \\
\midrule
TCLab         & Finetune  & Adam    & $0.9943 \pm 0.0078$ \\
              &           & L-BFGS  & $0.9998 \pm 0.0005$ \\
              &           & SEKF    & $0.9887 \pm 0.0171$ \\
              & Retrain   & Adam    & $0.8557 \pm 0.1274$ \\
              &           & L-BFGS  & $0.9305 \pm 0.0820$ \\
              &           & SEKF    & $0.7505 \pm 0.2451$ \\
\bottomrule
\end{tabular}
\end{table}

\begin{figure*}
\centering
\includegraphics[width=0.95\textwidth]{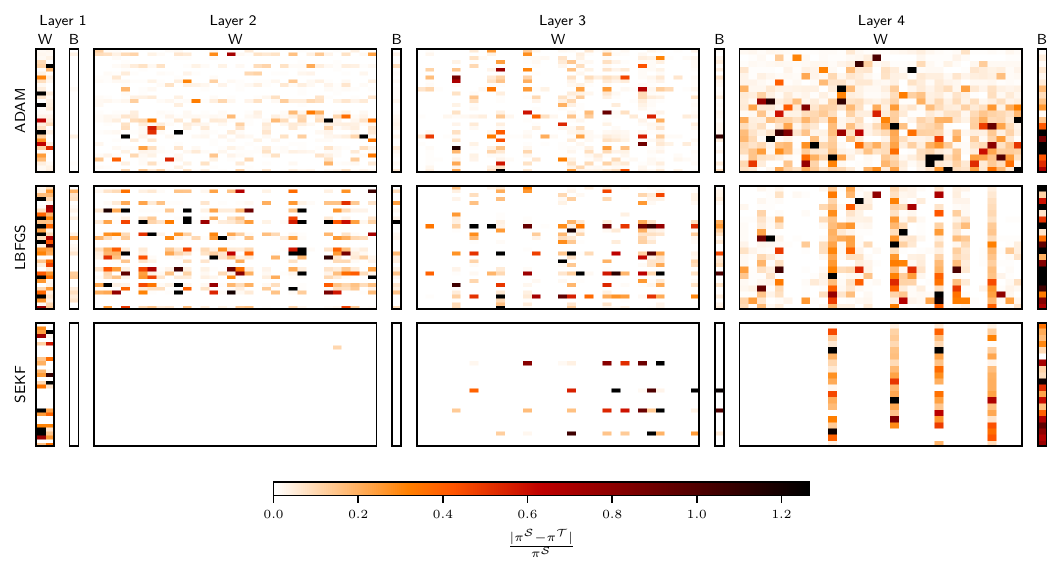}
\caption{Changes in model weights during transfer learning for the damped spring system when $c\scr\tgt = 0.9 c\scr\src$ and 1,000 target examples. Each subplot shows the distribution of weight changes in a specific layer of the neural network. Contrary to common heuristics in image classification, weight changes are distributed across all layers rather than concentrated in later layers. This suggests that transfer learning for dynamical systems requires holistic parameter adjustments.}
\label{fig:weight_changes_damped_spring}
\end{figure*}

Contrary to common heuristics in image classification transfer learning, Figure~\ref{fig:weight_changes_damped_spring} reveals that weight changes during finetuning distribute across all layers of the neural network rather than concentrating in later layers when finetuning using Adam or L-BFGS. In contrast, changes to model parameters are concentrated in a few parameters in the earlier and later layers when using the SEKF to finetune. Updates performed by the Adam algorithm tend to be smaller in magnitude, and more uniformly distributed across more parameters. In contrast, updates performed by L-BFGS tend to be larger in magnitude, and concentrated in fewer parameters. SEKF updates even more sparse, and restricted to just a few neurons.

Considered jointly, Figures~\ref{fig:cosine_similarity} and~\ref{fig:weight_changes_damped_spring} reveal that effective transfer learning for dynamical systems requires small but holistic parameter changes. Figure~\ref{fig:cosine_similarity} establishes that the \emph{magnitude} of change is small---finetuned parameters remain very close to the source---while Figure~\ref{fig:weight_changes_damped_spring} demonstrates that these small changes may often be distributed throughout the architecture. This combination challenges the standard computer vision transfer learning paradigm, where early layers encoding generic features are typically frozen while only later task-specific layers are adapted. For dynamical systems, the optimal strategy permits all layers to adapt while constraining the overall adaptation to remain near the source parameters.

The three optimizers achieve this distributed adaptation through distinct mechanisms visible in Figure~\ref{fig:weight_changes_damped_spring}. Adam produces small, uniformly distributed updates across a large number of parameters in every layer, reflecting its adaptive learning rate that scales updates inversely with gradient history. L-BFGS concentrates larger updates in fewer parameters, leveraging second-order curvature information to identify the most influential weights for reducing prediction error. SEKF exhibits the most selective behavior, restricting updates to specific neurons while leaving others unchanged; this sparsity arises from the subset selection mechanism that updates only parameters with the highest expected impact on prediction uncertainty. Despite these contrasting update patterns, all three optimizers achieve a final cosine similarity of greater than 0.98 when compared with source parameters (Table~\ref{tab:cosine_similarity}), suggesting that multiple paths through parameter space lead to similarly effective target models. This finding explains why optimizer choice does not significantly affect generalization performance (Section~\ref{sec:optimizer_effect})---the critical factor is maintaining proximity to the source model, which finetuning achieves regardless of the specific optimization trajectory.

This joint analysis also explains why finetuning substantially outperforms retraining: retrained models converge to parameter configurations with low cosine similarity to the source (Table~\ref{tab:cosine_similarity}), indicating they occupy different regions of parameter space entirely. Since successful transfer requires finding a specific ``nearby but distributed'' solution, retraining from random initialization has no mechanism to locate this favorable region, particularly when data is limited. The SEKF framework is well-suited to this setting because it treats source parameters as a Bayesian prior and computes full-network Jacobians, enabling coordinated updates across all layers while naturally constraining parameters to remain close to the well-generalized source model.

\begin{figure}[htbp]
\centering
\includegraphics[width=\columnwidth]{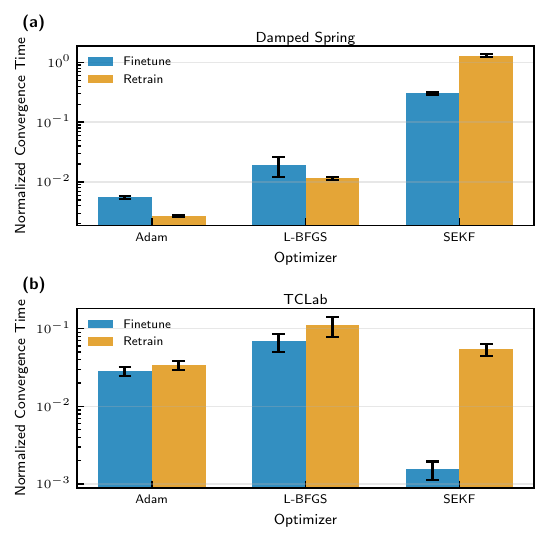}
\caption{Normalized convergence time by optimizer and initialization method. Left: Damped spring system. Right: TCLab system. SEKF exhibits the longest convergence times, especially when retraining. Finetuning reduces convergence time across all optimizers.}
\label{fig:convergence_time}
\end{figure}

Convergence time varies substantially across optimizers, as shown in Figure~\ref{fig:convergence_time}. SEKF requires significantly more computation than gradient-based methods, particularly when retraining from scratch. However, when finetuning, all optimizers converge more quickly due to the favorable initialization.

\subsection{Statistical Analysis}

Due to violations of standard ANOVA assumptions (non-normality, heteroscedasticity, and non-sphericity in the residuals), we employ two robust statistical methods: (1) Permutation ANOVA, a non-parametric approach that makes no distributional assumptions, and (2) Generalized Linear Models (GLM) with a Gamma distribution and log link, which appropriately models the positive, right-skewed distribution of loss values and convergence times as $\textrm{Normalized Metric} = \textrm{Parameter Initialization Effect}\times\textrm{Optimization Method Effect}\times\textrm{Datset Size Effect}$. The dependent variables analyzed are normalized test loss, train-test gap (a measure of overfitting), and normalized convergence time.

\subsubsection{Effect of Target Data Availability}

The amount of target domain data has the strongest and most consistent effect across all outcomes in both systems. Permutation ANOVA reveals highly significant main effects of data size on normalized test loss (damped spring: $F=8.29$, $p=2\times10^{-4}$; TCLab: $F=24.88$, $p=2\times10^{-4}$), train-test gap (damped spring: $F=6.02$, $p=2\times10^{-4}$; TCLab: $F=67.43$, $p=2\times10^{-4}$), and convergence time (damped spring: $F=14.19$, $p=2\times10^{-4}$; TCLab: $F=2.87$, $p=0.017$). The GLM coefficients quantify these effects: relative to the smallest dataset (10 samples for damped spring, 0.5 hours for TCLab), increasing to the largest dataset reduces log-normalized test loss by approximately 1.1 units (damped spring) and 1.2 units (TCLab), corresponding to roughly 3-fold improvements in prediction accuracy.

\subsubsection{Effect of Initialization Method}

Initialization method (finetuning vs.\ retraining) significantly affects model performance, though the magnitude differs between systems. For the damped spring system, Permutation ANOVA shows strong main effects on test loss ($F=6.58$, $p=2\times10^{-4}$), train-test gap ($F=4.73$, $p=2\times10^{-4}$), and convergence time ($F=64.04$, $p=2\times10^{-4}$). The GLM estimates that retraining increases log-normalized test loss by 1.89 units relative to finetuning ($p=1.88\times10^{-26}$), representing approximately a 6.6-fold increase in prediction error.

For the TCLab system, the effect of initialization on test loss is weaker: Permutation ANOVA yields $F=0.75$, $p=0.389$, though the GLM detects a marginally significant effect ($\beta=0.38$, $p=0.038$). However, initialization significantly affects generalization: retraining produces larger train-test gaps in both systems (damped spring: $\beta=1.08$, $p=4.50\times10^{-10}$; TCLab: $\beta=0.29$, $p=0.025$), confirming that finetuning provides implicit regularization against overfitting.

\subsubsection{Interaction Between Initialization and Data Size}%
\label{sec:interaction}
A critical finding emerges from the GLM interaction terms: the advantage of finetuning over retraining depends strongly on data availability. For the damped spring system, the interaction coefficients between retraining and data size are highly significant and negative (e.g., retrain $\times$ 1000 samples: $\beta=-1.88$, $p=6.56\times10^{-14}$), indicating that the performance gap between initialization methods narrows substantially as more target data becomes available. At the smallest data size (10 samples), retraining yields normalized test loss approximately 6.6$\times$ higher than finetuning; at 1000 samples, this gap shrinks to approximately 1.0$\times$---effectively equivalent performance.

The TCLab system shows a similar but weaker pattern. The interaction terms are not individually significant at $\alpha=0.05$, but the direction is consistent: finetuning's advantage diminishes with increasing data. This confirms that transfer learning provides the greatest benefit precisely when it is most needed---in data-scarce scenarios.

\subsubsection{Effect of Optimizer}%
\label{sec:optimizer_effect}
Optimizer choice (Adam, L-BFGS, SEKF) has context-dependent effects. For test loss, optimizer effects are not significant in the damped spring system (Permutation ANOVA: $F=1.05$, $p=0.394$) but are significant in TCLab ($F=6.46$, $p=0.002$), suggesting that optimizer selection becomes more important when dealing with noisy real-world data. The SEKF's advantage in the TCLab system likely reflects the benefit of explicit noise modeling in sim-to-real transfer: the measurement noise covariance $\bm{R}$ directly accounts for sensor noise present in physical data, while the process noise covariance $\bm{Q}$ controls how far parameters may deviate from the simulation-trained prior. In the damped spring case, where both source and target are simulated with controlled noise, this probabilistic machinery offers less benefit over gradient-based methods.

Critically, optimizer choice does \emph{not} significantly affect the train-test gap in either system (damped spring: $F=1.10$, $p=0.330$; TCLab: $F=0.32$, $p=0.739$). All three optimizers achieve statistically indistinguishable generalization performance despite following different optimization trajectories. This null result has practical significance: practitioners can select optimizers based on computational constraints without sacrificing generalization.

For convergence time, optimizer effects are highly significant in both systems (damped spring: $F=322.9$, $p=2\times10^{-4}$; TCLab: $F=9.62$, $p=2\times10^{-4}$). The GLM reveals that SEKF requires substantially more computation than gradient-based methods due to the matrix inversion performed. Relative to Adam, the SEKF increases log-normalized time by 2.14 units in the damped spring system ($p=5.01\times10^{-13}$), corresponding to approximately 8.5$\times$ longer training times. However, a significant interaction between SEKF and initialization method (damped spring: $\beta=3.42$, $p=9.99\times10^{-15}$; TCLab: $\beta=2.02$, $p=0.024$) indicates that SEKF's computational overhead is most pronounced when retraining from scratch; when finetuning, the relative disadvantage is reduced because parameter updates remain localized near the well-initialized source parameters.

\subsubsection{Summary of Statistical Findings}

Table~\ref{tab:statistical_summary} summarizes the key statistical findings. The evidence strongly supports the conclusions that (1) target data availability is the dominant factor affecting all outcomes, (2) finetuning outperforms retraining, especially when data is limited, (3) finetuning provides regularization benefits reflected in smaller train-test gaps, and (4) optimizer choice affects computational efficiency but not generalization quality.

\begin{table*}[htbp]
\centering
\caption{Summary of Permutation ANOVA results ($F$-statistics and $p$-values) for main effects across both benchmark systems. Significant effects ($p < 0.05$) are shown in bold.}
\label{tab:statistical_summary}
\small
\begin{tabular}{llcccc}
\toprule
& & \multicolumn{2}{c}{Damped Spring} & \multicolumn{2}{c}{TCLab} \\
\cmidrule(lr){3-4} \cmidrule(lr){5-6}
Outcome & Factor & $F$ & $p$ & $F$ & $p$ \\
\midrule
\multirow{3}{*}{Test Loss}
& Data Size & 8.29 & \textbf{$2\times10^{-4}$} & 24.88 & \textbf{$2\times10^{-4}$} \\
& Init Method & 6.58 & \textbf{$2\times10^{-4}$} & 0.75 & 0.389 \\
& Optimizer & 1.05 & 0.394 & 6.46 & \textbf{0.002} \\
\midrule
\multirow{3}{*}{Train-Test Gap}
& Data Size & 6.02 & \textbf{$2\times10^{-4}$} & 67.43 & \textbf{$2\times10^{-4}$} \\
& Init Method & 4.73 & \textbf{$2\times10^{-4}$} & 7.87 & \textbf{0.007} \\
& Optimizer & 1.10 & 0.330 & 0.32 & 0.739 \\
\midrule
\multirow{3}{*}{Conv.\ Time}
& Data Size & 14.19 & \textbf{$2\times10^{-4}$} & 2.87 & \textbf{0.017} \\
& Init Method & 64.04 & \textbf{$2\times10^{-4}$} & 6.11 & \textbf{0.008} \\
& Optimizer & 322.94 & \textbf{$2\times10^{-4}$} & 9.62 & \textbf{$2\times10^{-4}$} \\
\bottomrule
\end{tabular}
\end{table*}

\section{Conclusion}
\label{sec:conclusion}

This work investigated transfer learning for neural network models of dynamical systems, demonstrating that pre-trained source models can be efficiently adapted to related target systems using severely limited data. The experimental evidence from both simulated (damped spring-mass) and physical (Temperature Control Laboratory) systems supports several conclusions with implications for practitioners and researchers working on data-driven modeling of dynamical systems.

\textbf{Finetuning substantially outperforms retraining from scratch.} Across all experimental conditions, initializing from pre-trained source parameters yields significantly lower test error than random initialization. This advantage is most pronounced when target data is limited---with as few as 10 samples or 0.5 hours of operational data, finetuning achieves performance that retraining cannot match even with substantially more data. For industrial applications where data collection is constrained by cost, safety, or time, this finding suggests that investing in high-quality source models and leveraging transfer learning provides greater returns than attempting to train target-specific models from scratch.

\textbf{Effective transfer requires only small parameter perturbations.} Finetuned models maintain cosine similarity exceeding 0.99 with source parameters, demonstrating that successful adaptation occurs within a small neighborhood of weight space. This proximity validates the underlying assumption that functionally similar dynamical systems---those sharing governing equation structure and overlapping operating domains---can be modeled by neural networks with similar parameters. The practical implication is that transfer learning for dynamical systems is well-suited to parameter-efficient methods that constrain updates to remain near the source model, reducing both computational cost and overfitting risk.

\textbf{Finetuning provides implicit regularization against overfitting.} When training data is limited, overfitting poses a fundamental challenge. The experimental results demonstrate that finetuning exhibits smaller train-test gaps than retraining, indicating superior generalization. This regularization effect can be understood through the Bayesian lens: source parameters define an informative prior $p(\param) = \mathcal{N}(\param\scr\src, \cov_0)$ that biases optimization toward regions of parameter space with favorable generalization properties. When target data is scarce, the posterior remains close to this prior because observations lack sufficient evidence to shift the distribution substantially. Retraining from random initialization corresponds to an uninformative prior that offers no such guidance, explaining its inferior generalization. For practitioners, this suggests that transfer learning offers benefits beyond reduced data requirements---it also improves model reliability on unseen operating conditions.

\textbf{Optimizer choice affects efficiency but not generalization.} While Adam, L-BFGS, and SEKF differ substantially in convergence time, they achieve statistically indistinguishable generalization performance as measured by train-test gap. This null result has practical significance: practitioners can select optimizers based on computational constraints, available infrastructure, or convergence speed without sacrificing model quality. The SEKF, despite its Bayesian formulation and theoretical regularization properties, does not provide measurable generalization advantages over simpler gradient-based methods in the experimental settings studied. However, the SEKF's sequential processing enables online adaptation during normal system operation, where convergence time is less critical because the system continues to operate while the model adapts. In contrast, gradient-based methods require assembling batch datasets, implying discrete data collection and model redeployment cycles. The operational continuity afforded by online adaptation may outweigh the SEKF's slower convergence in practical deployments.

\textbf{Effective transfer requires small but distributed parameter changes.} Unlike transfer learning for image classification, where early layers encoding general features are typically frozen while later task-specific layers are adapted, dynamical system transfer learning exhibits distributed parameter modifications across all layers while maintaining high cosine similarity with source parameters. This combination---small magnitude changes distributed throughout the architecture---suggests that the hierarchical feature abstraction paradigm underlying layer-freezing heuristics in computer vision does not directly transfer to regression problems in dynamical systems. Notably, the three optimizers achieve this distributed adaptation through distinct mechanisms: Adam produces small, uniform updates across many parameters; L-BFGS concentrates larger updates in fewer influential weights; and SEKF exhibits the most selective behavior, restricting updates to specific neurons. Despite these contrasting update patterns, all optimizers achieve comparable generalization, indicating that multiple paths through parameter space lead to similarly effective target models provided they remain near the source. Practitioners should therefore adapt all network parameters rather than selectively freezing layers when applying transfer learning to system identification tasks.

\textbf{Implications for modeling dynamical systems with limited data.} The results presented here suggest a practical workflow for developing neural network models of dynamical systems in data-scarce environments. First, train a high-quality source model using abundant data from a related system---whether simulated, historical, or from similar equipment. Second, collect minimal target data sufficient to capture the essential differences between source and target systems. Third, finetune the source model on target data using standard gradient-based optimization, which provides computational efficiency without sacrificing generalization. This approach enables accurate modeling with as little as 1\% of the data required for training from scratch, substantially lowering barriers to deploying data-driven methods in settings where comprehensive data collection is infeasible such as industrial, automotive, healthcare etc.

\textbf{Limitations and future directions.} The systems studied here satisfy the assumptions of functional similarity and overlapping operating domains. Transfer learning performance when these assumptions are violated---for example, when source and target systems have qualitatively different dynamics, or when the same model is deployed in significantly different conditions---remains to be characterized. Additionally, the neural network architectures employed are relatively small; scaling behavior to larger models and more complex systems warrants investigation. Future work should also explore adaptive methods that automatically determine the appropriate degree of parameter regularization based on the similarity between source and target systems, potentially enabling robust transfer even when system relationships are uncertain. The SEKF process model (Eq.~\eqref{eq:process_model}) already treats parameters as a random walk---an integrator driven by white noise---suggesting a natural extension to scenarios where physical parameters are non-constant but slowly varying.

\section*{Acknowledgments}

The authors would like to thank the industrial sponsors of the Texas Wisconsin California Control Consortium (TWCCC) for their support of this research.

\section*{Data and Code Availability}

All code necessary to reproduce the results in this paper is available at \url{https://github.com/joshuaeh/Subset-Extended-Kalman-Filter/tree/transfer}. No new data was generated for this study; all data used is either simulated or publicly available.

\printcredits

\bibliographystyle{unsrtnat}

\appendix
\onecolumn
\section{Experimental Details}
\label{app:exp_details}

\subsection{Damped Spring-Mass System}
\label{app:subsec:spring_mass_details}

The Damped Spring-Mass system is governed by the second-order ordinary differential equation:

\begin{equation}
    m \frac{d^2 x}{dt^2} + c \frac{dx}{dt} + k x = u,
\end{equation}

where \( x \) is the position of the mass, \( m \) is the mass, \( c \) is the damping coefficient, \( k \) is the spring constant, and \( u \) is the external force applied to the system. Source System parameters are listed in Table~\ref{tab:source_spring_mass_params}. The source NN model is trained on 100,000 datapoints generated by simulating the system with uniformly sampled random initial positions and velocities over a time horizon of 20 seconds. The MLP model consists of 2 hidden layers with 32 neurons each and sigmoid activation functions. The model is trained using the Adam optimizer with hyperparameters listed in Table~\ref{tab:source_spring_hyperparameters} tuned with the ASHA algorithm using the \texttt{ray} python library.

\begin{table}
\caption{Source Damped Spring-Mass System Parameters}
\label{tab:source_spring_mass_params}
\centering
\begin{tabular}{lcl}
\toprule
\textbf{Parameter} & \textbf{Value} & \textbf{Units}\\
\midrule
Mass, \( m \) & 1.0 & kg \\
Damping Coefficient, \( c \) & 0.5 & Ns/m \\
Spring Constant, \( k \) & 1.0 & N/m \\
Force, \( u \) & 0 & N \\
\midrule
Initial Position \(x_0 \) & \text{Uniform}(-5, 5) & m \\
Initial Velocity \(\dot{x} \)& \text{Uniform}(-5, 5) & m/s \\
\bottomrule
\end{tabular}
\end{table}

\begin{table}
\caption{Source Damped Spring-Mass System NN Hyperparameters}
\label{tab:source_spring_hyperparameters}
\centering
\begin{tabular}{ll}
\toprule
\textbf{Hyperparameter} & \textbf{Values} \\
\midrule
Learning Rate & $10^{-6} \textnormal{--} 10^{-1}$\\
Batch Size & 16,32,64,128 \\
LR Patience & 10, 20, 50, 100 \\
LR Factor & 0.1 --  0.9  \\
\midrule
Maximum Epochs & 1000 \\
Num. Samples & 100 \\
Max. Concurrent Samples & 4 \\
\bottomrule
\end{tabular}
\end{table}

The 8 target systems differ from the source system by changing one of the physical parameters of the system by $\pm 10\%$ ($\pm 1$ N in the case of force $u$). Each system is similarly simulated to generate 100,000 input-output pairs for training the target models separated into 10 replicates of 10,000 pairs each. Within each target dataset, 9,000 pairs are held out as a test dataset to approximate the generalization error of the target models. Training and validation datasets are created by selecting the first 10, 50, 100, 500, and 1,000 datapoints from the remaining 1,000 pairs and splitting them into 90\% training and 10\% validation sets.

\begin{table}
\centering
\caption{Summary of all Damped Spring-Mass Trials}
\label{tab:spring_mass_experiments}
\begin{tabular}{rl}
\toprule
2 & Parameter Initialization Methods: Finetune, Retrain \\
3 & Optimization Methods: SEKF, Adam, LBFGS \\
8 & Target Systems: $\pm 10\%$ change in $m$, $c$, $k$, $u$ \\
5 & Training Dataset Sizes: 10, 50, 100, 500, 1,000 \\
10 & Replicates per Target System and Dataset Size \\
\midrule
2,400 & Total Trials \\
\bottomrule
\end{tabular}
\end{table}

\begin{table}
\centering
\caption{Transfer Learning Hyperparameters}
\label{tab:spring_mass_tl_hyperparameters}
\begin{tabular}{lll}
\toprule
All & Mini-batches per Epoch & 50 \\
{} & Max. Epochs & 100 \\
\midrule
Adam & Learning Rate & $10^{-6} \textnormal{--} 10^{-1}$\\
{} & Mini-batch Size & $1$--$8,000$ \\
{} & LR Patience & 10\\
{} & LR Factor & 0.1 --  0.9  \\
\midrule
LBFGS & Learning Rate & $10^{-6} \textnormal{--} 1$\\
{} & Mini-batch Size & 1--8,000 \\
{} & History Size & 10 \\
{} & Max. Line Searches & 10 \\
\midrule
SEKF & Measurement Noise $R$ & 0.01 \\
{} & Process Noise $Q$ & $10^{-6}, 10^{-4}, 10^{-2}, 10^{-1}$ \\
{} & Initial Covariance $P_0$ & 0.01, 1.0, 10, 100 \\
{} & Mini-batch Size & 1, 2, 4, 8 \\
\bottomrule
\end{tabular}
\end{table}

\subsection{Temperature Control Lab}
\label{app:subsec:tclab_details}

\begin{figure}[htbp]
\centering
\includegraphics[width=0.6\columnwidth]{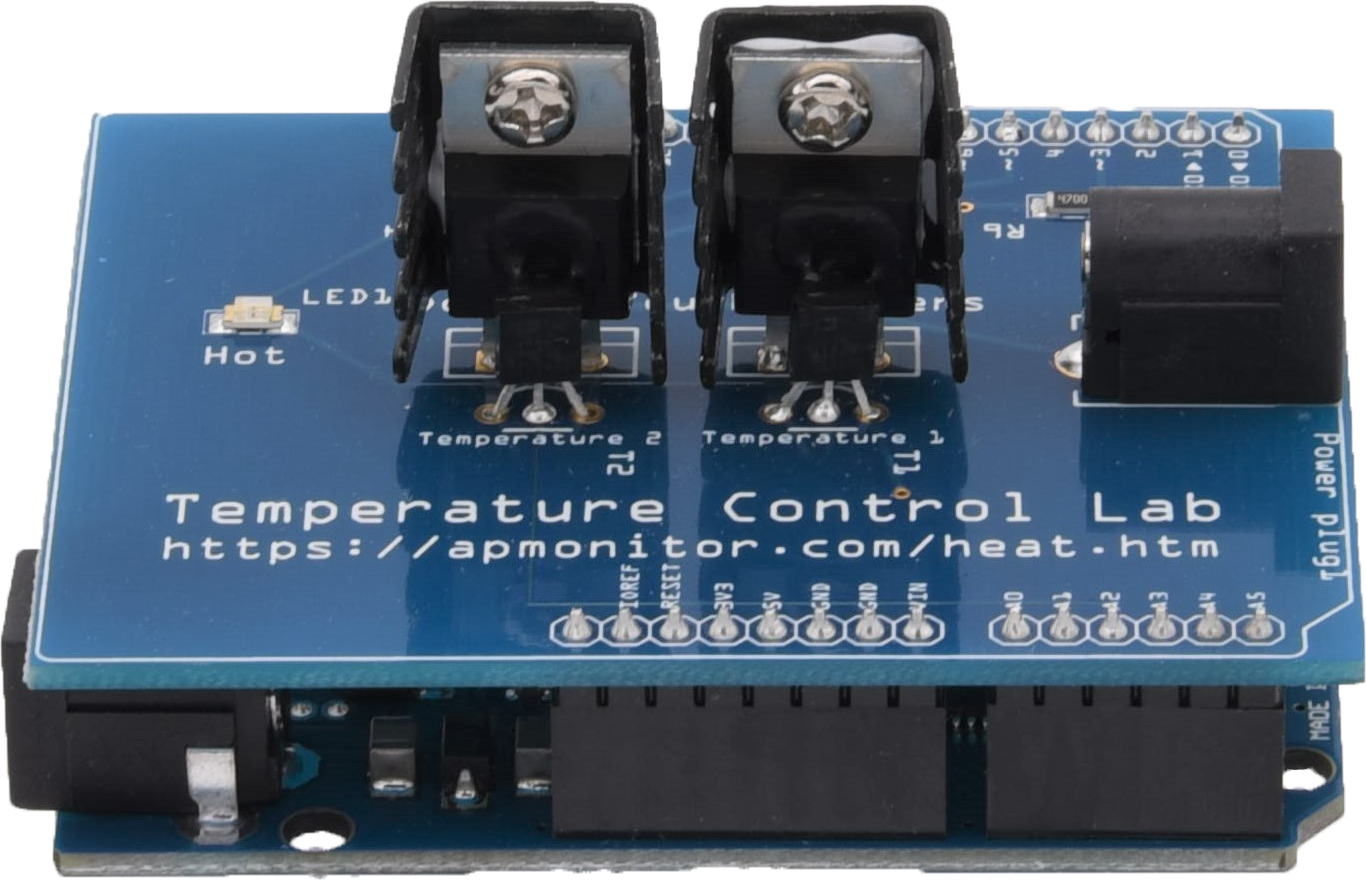}
\caption{The Temperature Control Lab (TCLab) hardware platform. Image courtesy of APMonitor.com.}
\label{fig:tclab_hardware}
\end{figure}

The Temperature Control Lab (TCLab) is a pocket-sized experimental platform consisting of two heaters and two temperature sensors mounted on an Arduino-based board, designed for teaching and research in process dynamics and control~\cite{park2020benchmark, de2020introducing}. Figure~\ref{fig:tclab_hardware} shows the physical device. We use a second-order physics-based model of the TCLab system based on energy balances for each heater~\cite{park2020benchmark}.

The governing equations for the two heater temperatures are coupled nonlinear energy balances:
\begin{align}
    m c_p \frac{dT_1}{dt} &= UA(T_\infty - T_1) + \epsilon\sigma A(T_\infty^4 - T_1^4) + Q_{C12} + Q_{R12} + \alpha_1 Q_1 \label{eq:tclab_T1}\\
    m c_p \frac{dT_2}{dt} &= UA(T_\infty - T_2) + \epsilon\sigma A(T_\infty^4 - T_2^4) - Q_{C12} - Q_{R12} + \alpha_2 Q_2 \label{eq:tclab_T2}
\end{align}
where $T_1$ and $T_2$ are the heater temperatures, $T_\infty$ is the ambient temperature, and $Q_1$ and $Q_2$ are the heater power inputs (0--100\%). The inter-heater heat transfer comprises convective and radiative components:
\begin{align}
    Q_{C12} &= U A_s (T_2 - T_1) \label{eq:tclab_Qc}\\
    Q_{R12} &= \epsilon \sigma A_s (T_2^4 - T_1^4) \label{eq:tclab_Qr}
\end{align}

Table~\ref{tab:tclab_params} lists the model parameters used for simulation.

\begin{table}[htbp]
\centering
\caption{TCLab Model Parameters}
\label{tab:tclab_params}
\begin{tabular}{lcc}
\toprule
\textbf{Parameter} & \textbf{Value} & \textbf{Units} \\
\midrule
Mass, $m$ & 0.004 & kg \\
Heat Capacity, $c_p$ & 500 & J/(kg$\cdot$K) \\
Heat Transfer Coefficient, $U$ & 10 & W/(m$^2\cdot$K) \\
Exposed Surface Area, $A$ & $1.0 \times 10^{-3}$ & m$^2$ \\
Inter-heater Surface Area, $A_s$ & $2.0 \times 10^{-4}$ & m$^2$ \\
Emissivity, $\epsilon$ & 0.9 & -- \\
Stefan-Boltzmann Constant, $\sigma$ & $5.67 \times 10^{-8}$ & W/(m$^2\cdot$K$^4$) \\
Heater 1 Factor, $\alpha_1$ & 0.01 & W/\% \\
Heater 2 Factor, $\alpha_2$ & 0.0075 & W/\% \\
Ambient Temperature, $T_\infty$ & 296.15 (23$^\circ$C) & K \\
\bottomrule
\end{tabular}
\end{table}

The source system was simulated for one year with a measurement frequency of 10 seconds, generating approximately 3.15 million data points. Heater power inputs were generated using a pseudo-random scheme: at random intervals of 1--10 minutes, each heater was assigned a uniformly sampled power setting (0--100\%), with a 50\% probability of being set to zero to encourage exploration of cooling dynamics.

The neural network model predicts both heater temperatures over a 10-minute (60 sample) prediction horizon given initial temperatures and heater power inputs across the horizon. A Neural ODE model is employed to capture the dynamics of the system. An architecture with two hidden layers is utilized with 32 neurons each layer and sigmoid activation.

The target system represents a physical TCLab unit, with real-world disturbances such as sensor noise and unmodeled dynamics. The target system was operated for one week, collecting data at 10-second intervals. Heater power inputs followed the same pseudo-random scheme as the source system to ensure comparable excitation. Then, each day was used as a separate training dataset with the final two days held out for testing. Within each training dataset, training and validation sets were created by selecting the first 0.5 hours, 1 hour, 4 hours, 12 hours, and 24 hours of data and splitting them into 90\% training and 10\% validation sets.

\begin{table}
\caption{Summary of all TCLab Trials}
\label{tab:cstr_trials}
\centering
\begin{tabular}{rl}
\toprule
2 & Parameter Initialization Methods: Finetune, Retrain \\
3 & Optimization Methods: SEKF, Adam, LBFGS \\
5 & Training Dataset Sizes: 0.5 hours, 1 hour, 4 hours, 12 hours, 24 hours \\
5 & Replicates \\
\midrule
150 & Total Trials \\
\bottomrule
\end{tabular}

\end{table}

\section{Supplementary Data}
\label{app:supp_data}

This appendix provides additional visualizations that complement the main text figures. These include full scatter plots showing individual trial results, detailed breakdowns by system and replicate, and diagnostic plots for statistical model validation.

\subsection{Example Predictions}
\begin{figure*}[htbp]
\centering
\includegraphics[width=0.95\textwidth]{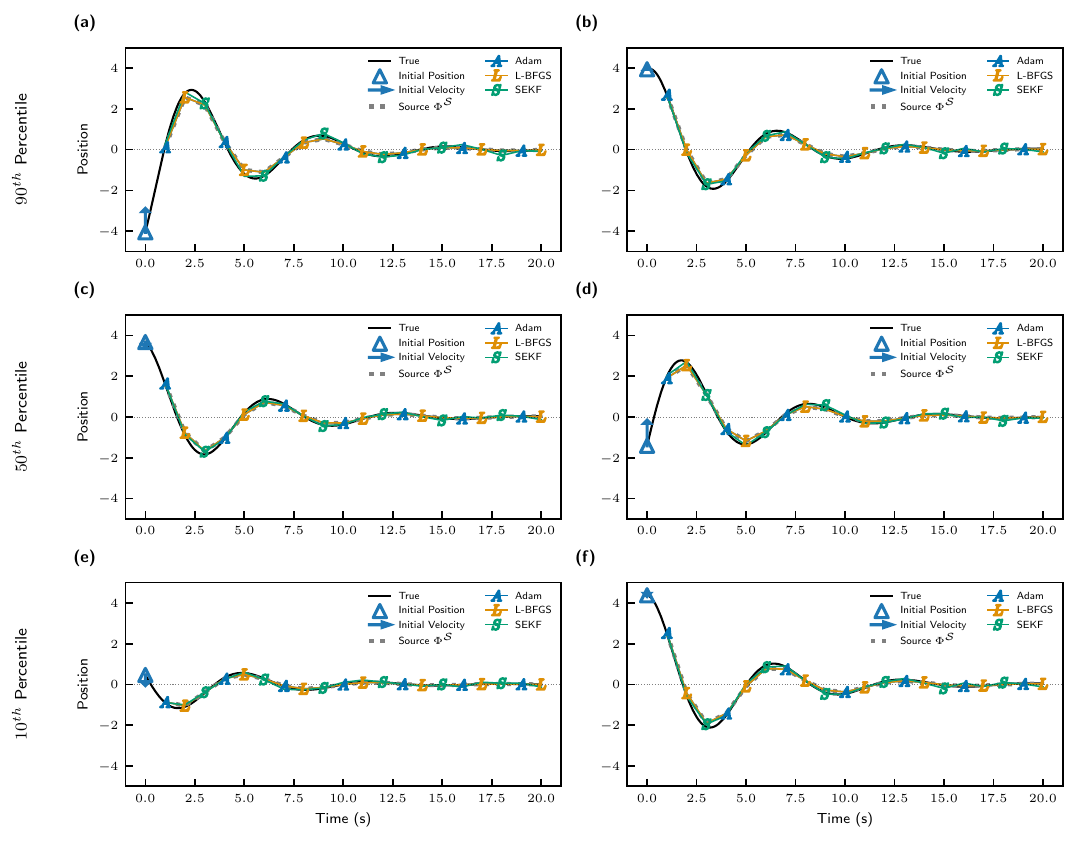}
\caption{Example prediction trajectories for the damped spring system at the 90th, 50th, and 10th percentiles of test loss for the source model. Target system has $c\scr\tgt = 0.45$ (10\% reduction from source). Finetuned models (middle row) maintain accuracy comparable to the source model, while retrained models (bottom row) exhibit larger prediction errors, especially in worst-case scenarios.}
\label{fig:spring_predictions}
\end{figure*}

\begin{figure*}[htbp]
\centering
\includegraphics[width=0.95\textwidth]{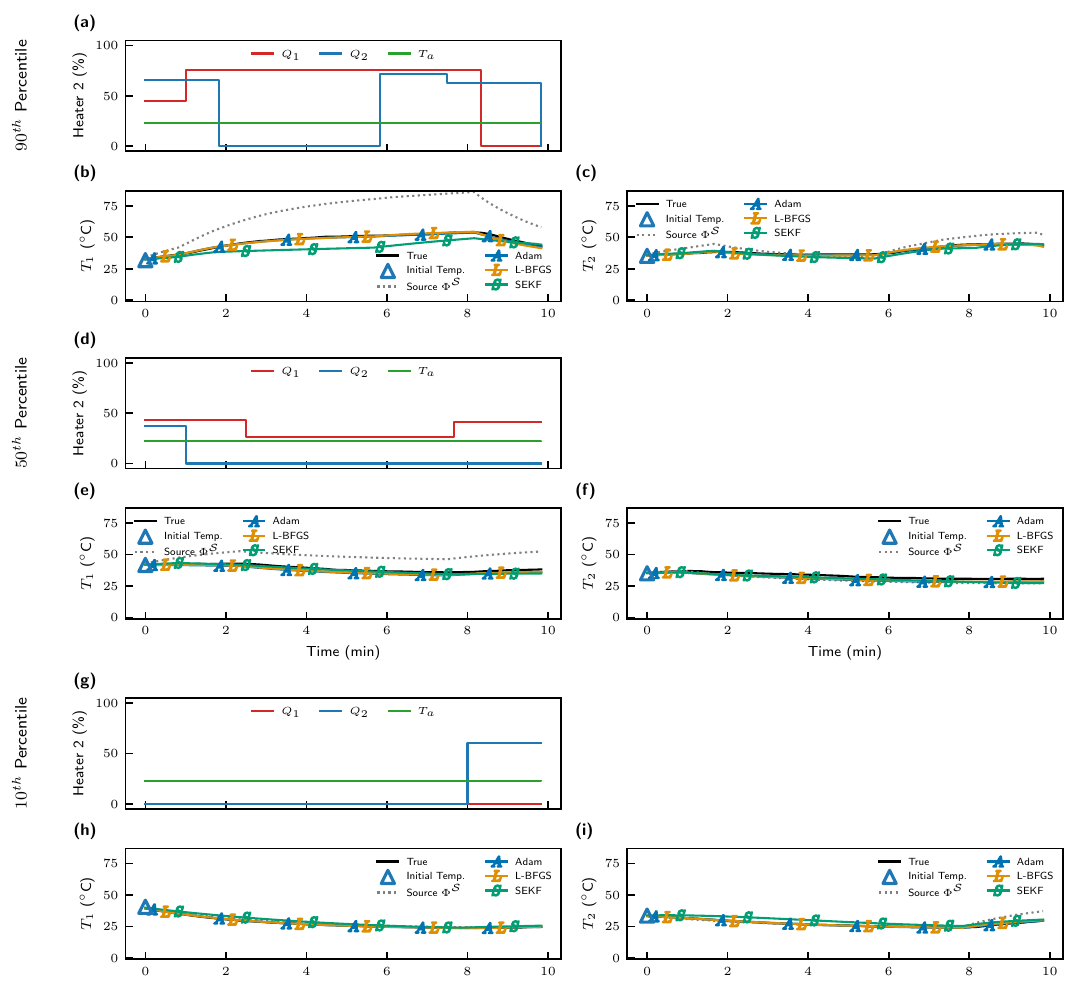}
\caption{Example prediction trajectories for the TCLab system at the 90th, 50th, and 10th percentiles of test loss for the source model. Transfer learning adapts a simulation-trained model to physical hardware using 4 hours of experimental data. Finetuned models (middle row) accurately predict both heater temperatures, while retrained models (bottom row) show degraded performance, particularly for the second heater.}
\label{fig:tclab_predictions}
\end{figure*}

Example prediction trajectories from both the damped spring and TCLab systems are selected by choosing specific examples at the 90th, 50th, and 10th percentiles of test loss for each initialization method. Figure~\ref{fig:spring_predictions} shows damped spring trajectories, comparing source model predictions (trained on $c\scr\src = 0.5$) with finetuned and retrained target models (adapting to $c\scr\tgt = 0.45$) using 100 training samples. The finetuned model closely tracks the target system even in the worst-case scenario (90th percentile), while the retrained model exhibits substantial deviation. Figure~\ref{fig:tclab_predictions} presents analogous results for the TCLab system, demonstrating transfer from simulation to physical hardware using 4 hours of real-world data. These visualizations confirm that finetuning maintains predictive accuracy across the distribution of outcomes, whereas retraining produces more variable and often degraded performance.

\subsection{Full Test Loss Results}

\begin{figure*}[htbp]
\centering
\includegraphics[width=0.95\textwidth]{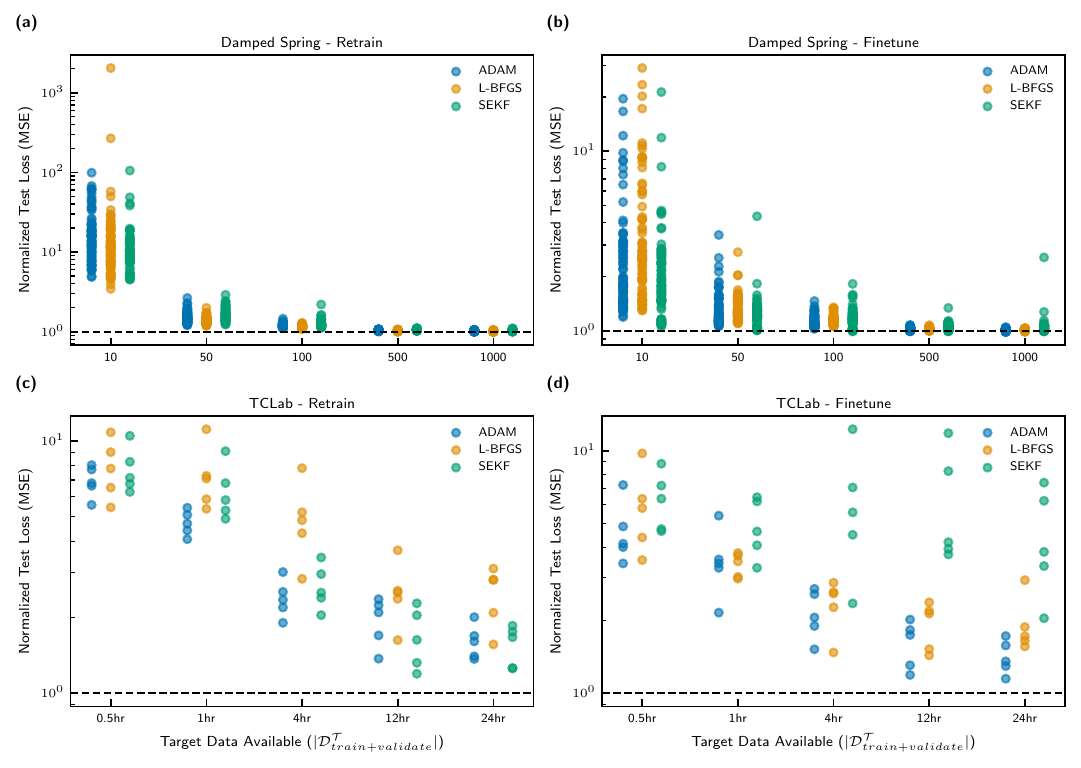}
\caption{Complete scatter plots of normalized test loss for all individual trials across both experimental systems. Each point represents a single trial, colored by initialization method (blue: finetune, orange: retrain). Panels are organized by target data size.}
\label{fig:supp_test_loss_full}
\end{figure*}

Figure~\ref{fig:supp_test_loss_full} presents complete strip plots of normalized test loss for all individual trials across both experimental systems. Note that strip plots represent the target data available as a categorical variable and thus the horizontal variance within each condition is presented to characterize differences between optimizers, not to suggest that one optimizer has more data than another. Each point represents a single trial, colored by initialization method (blue: finetune, orange: retrain). Panels are organized by target data size to illustrate the distribution of outcomes within each condition.

\begin{figure*}[htbp]
\centering
\includegraphics[width=0.95\textwidth]{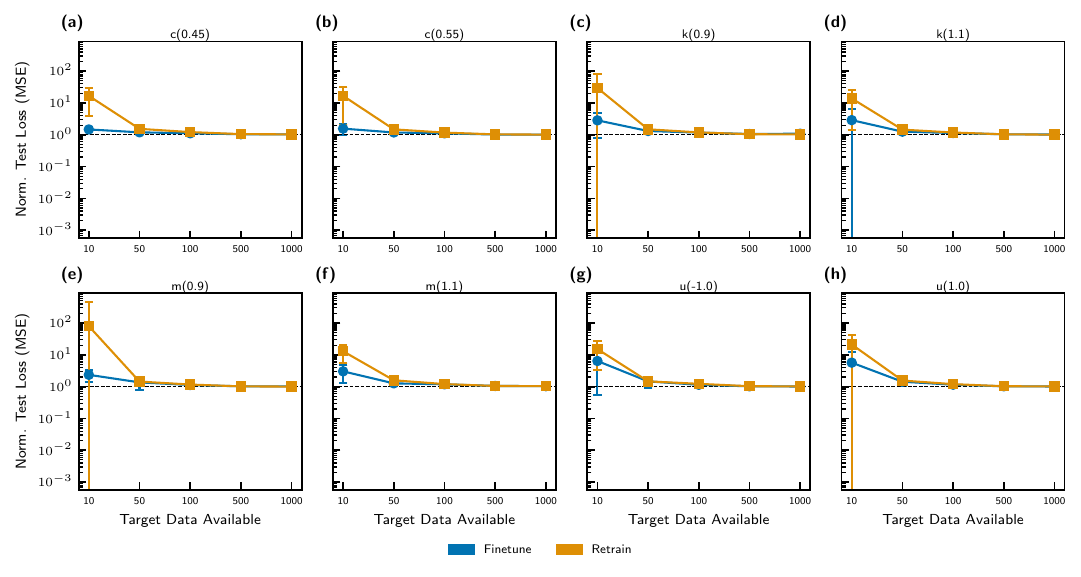}
\caption{Normalized test loss broken down by target system for the damped spring experiments. Each of the 8 target systems (varying $m$, $c$, $k$, $u$ by $\pm 10\%$) is shown separately to illustrate consistency of transfer learning benefits across different parameter perturbations.}
\label{fig:supp_test_loss_by_system}
\end{figure*}

\subsection{Train-Test Gap Details}

\begin{figure*}[htbp]
\centering
\includegraphics[width=0.95\textwidth]{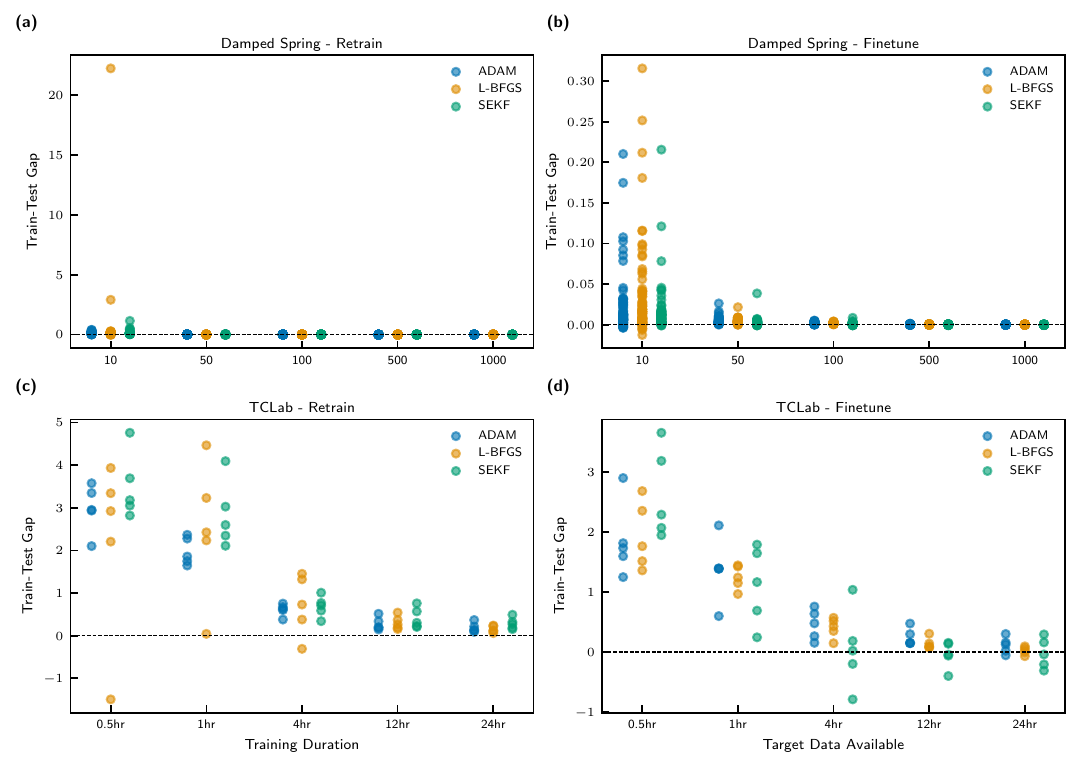}
\caption{Complete scatter plots of train-test gap (difference between training and test loss) for all trials. Lower values indicate better generalization. Finetuning consistently shows smaller gaps, particularly at low data sizes.}
\label{fig:supp_gap_full}
\end{figure*}

Figure~\ref{fig:supp_gap_full} presents complete scatter plots of the train-test gap for all individual trials, providing a detailed view of generalization performance across experimental conditions.

\subsection{Convergence Time Analysis}

\begin{figure*}[htbp]
\centering
\includegraphics[width=0.9\textwidth]{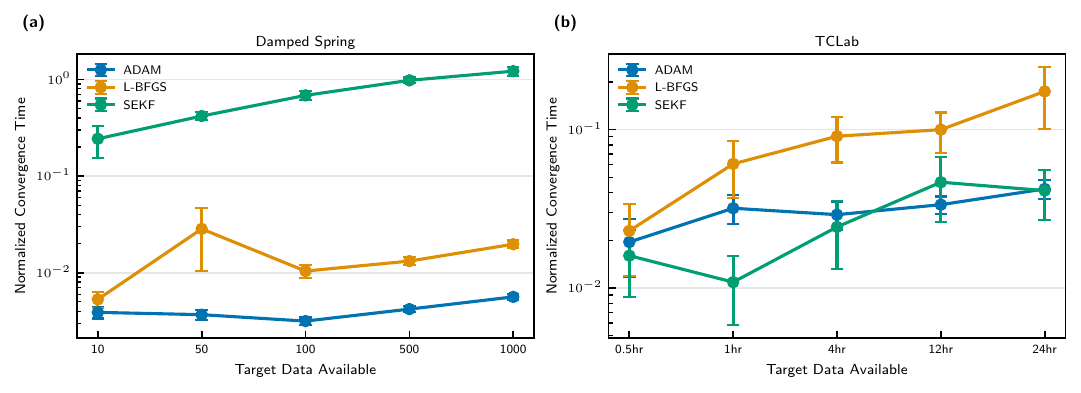}
\caption{Normalized convergence time broken down by target data size for each optimizer. This visualization reveals how optimizer efficiency scales with data availability and whether the relative performance of optimizers changes across data regimes.}
\label{fig:supp_convergence_by_data}
\end{figure*}

Figure~\ref{fig:supp_convergence_by_data} shows normalized convergence time broken down by target data size for each optimizer, illustrating how computational efficiency varies across data regimes and optimization methods.

\subsection{Statistical Model Diagnostics}

\section{Statistical Analysis Details}
\label{app:statistical_details}

This appendix presents the complete GLM coefficient tables from the Gamma regression models used in the statistical analysis. Tables include only coefficients with $p < 0.05$.

\subsection{Damped Spring System}

Tables~\ref{tab:glm_spring_test_loss}--\ref{tab:glm_spring_time} present the complete GLM coefficient estimates for the damped spring system, covering normalized test loss, train-test gap, and convergence time outcomes respectively.

\begin{table}[htbp]
\centering
\caption{GLM (Gamma) significant coefficients for normalized test loss---Damped Spring. Reference levels: Adam optimizer, Finetune initialization, 10 training samples.}
\label{tab:glm_spring_test_loss}
\small
\begin{tabular}{lrrrr}
\toprule
Effect & Coef. & Std.Err. & $z$ & $p$ \\
\midrule
Intercept & 1.120 & 0.125 & 8.93 & $4.11\times10^{-19}$ \\
L-BFGS & 0.352 & 0.177 & 1.99 & $4.70\times10^{-2}$ \\
Retrain & 1.887 & 0.177 & 10.64 & $1.88\times10^{-26}$ \\
50 Samples & $-$0.844 & 0.177 & $-$4.76 & $1.94\times10^{-6}$ \\
100 Samples & $-$1.005 & 0.177 & $-$5.67 & $1.47\times10^{-8}$ \\
500 Samples & $-$1.093 & 0.177 & $-$6.17 & $7.00\times10^{-10}$ \\
1000 Samples & $-$1.108 & 0.177 & $-$6.25 & $4.17\times10^{-10}$ \\
Retrain $\times$ 50 Samples & $-$1.749 & 0.251 & $-$6.98 & $3.04\times10^{-12}$ \\
Retrain $\times$ 100 Samples & $-$1.825 & 0.251 & $-$7.28 & $3.34\times10^{-13}$ \\
Retrain $\times$ 500 Samples & $-$1.878 & 0.251 & $-$7.49 & $6.99\times10^{-14}$ \\
Retrain $\times$ 1000 Samples & $-$1.880 & 0.251 & $-$7.50 & $6.56\times10^{-14}$ \\
\bottomrule
\end{tabular}
\end{table}

\begin{table}[htbp]
\centering
\caption{GLM (Gamma) significant coefficients for train-test gap---Damped Spring. Reference levels: Adam optimizer, Finetune initialization, 10 training samples.}
\label{tab:glm_spring_gap}
\small
\begin{tabular}{lrrrr}
\toprule
Effect & Coef. & Std.Err. & $z$ & $p$ \\
\midrule
Intercept & $-$2.929 & 0.122 & $-$23.99 & $3.85\times10^{-127}$ \\
Retrain & 1.077 & 0.173 & 6.24 & $4.50\times10^{-10}$ \\
50 Samples & $-$0.366 & 0.173 & $-$2.12 & $3.40\times10^{-2}$ \\
100 Samples & $-$0.431 & 0.173 & $-$2.49 & $1.27\times10^{-2}$ \\
500 Samples & $-$0.475 & 0.173 & $-$2.75 & $6.00\times10^{-3}$ \\
1000 Samples & $-$0.481 & 0.173 & $-$2.78 & $5.40\times10^{-3}$ \\
L-BFGS $\times$ Retrain & 0.815 & 0.244 & 3.34 & $8.50\times10^{-4}$ \\
Retrain $\times$ 50 Samples & $-$1.011 & 0.244 & $-$4.14 & $3.50\times10^{-5}$ \\
Retrain $\times$ 100 Samples & $-$1.048 & 0.244 & $-$4.29 & $1.76\times10^{-5}$ \\
Retrain $\times$ 500 Samples & $-$1.075 & 0.244 & $-$4.40 & $1.08\times10^{-5}$ \\
Retrain $\times$ 1000 Samples & $-$1.077 & 0.244 & $-$4.41 & $1.04\times10^{-5}$ \\
L-BFGS $\times$ Retrain $\times$ 50 Samples & $-$0.844 & 0.345 & $-$2.45 & $1.45\times10^{-2}$ \\
L-BFGS $\times$ Retrain $\times$ 100 Samples & $-$0.836 & 0.345 & $-$2.42 & $1.55\times10^{-2}$ \\
L-BFGS $\times$ Retrain $\times$ 500 Samples & $-$0.818 & 0.345 & $-$2.37 & $1.78\times10^{-2}$ \\
L-BFGS $\times$ Retrain $\times$ 1000 Samples & $-$0.816 & 0.345 & $-$2.36 & $1.82\times10^{-2}$ \\
\bottomrule
\end{tabular}
\end{table}

\begin{table}[htbp]
\centering
\caption{GLM (Gamma) significant coefficients for normalized convergence time---Damped Spring. Reference levels: Adam optimizer, Finetune initialization, 10 training samples.}
\label{tab:glm_spring_time}
\small
\begin{tabular}{lrrrr}
\toprule
Effect & Coef. & Std.Err. & $z$ & $p$ \\
\midrule
Intercept & $-$5.161 & 0.209 & $-$24.67 & $2.33\times10^{-134}$ \\
SEKF & 2.138 & 0.296 & 7.23 & $5.01\times10^{-13}$ \\
Retrain & $-$1.036 & 0.296 & $-$3.50 & $4.65\times10^{-4}$ \\
SEKF $\times$ Retrain & 3.418 & 0.442 & 7.74 & $9.99\times10^{-15}$ \\
L-BFGS $\times$ 50 Samples & 2.036 & 0.418 & 4.87 & $1.14\times10^{-6}$ \\
SEKF $\times$ 50 Samples & 1.543 & 0.418 & 3.69 & $2.26\times10^{-4}$ \\
SEKF $\times$ 100 Samples & 2.108 & 0.418 & 5.04 & $4.72\times10^{-7}$ \\
SEKF $\times$ 500 Samples & 2.336 & 0.418 & 5.58 & $2.36\times10^{-8}$ \\
SEKF $\times$ 1000 Samples & 2.126 & 0.418 & 5.08 & $3.77\times10^{-7}$ \\
L-BFGS $\times$ Retrain $\times$ 50 Samples & $-$1.176 & 0.592 & $-$1.99 & $4.69\times10^{-2}$ \\
SEKF $\times$ Retrain $\times$ 50 Samples & $-$1.286 & 0.624 & $-$2.06 & $3.95\times10^{-2}$ \\
SEKF $\times$ Retrain $\times$ 100 Samples & $-$1.247 & 0.625 & $-$1.99 & $4.62\times10^{-2}$ \\
SEKF $\times$ Retrain $\times$ 500 Samples & $-$1.526 & 0.625 & $-$2.44 & $1.47\times10^{-2}$ \\
SEKF $\times$ Retrain $\times$ 1000 Samples & $-$1.381 & 0.625 & $-$2.21 & $2.73\times10^{-2}$ \\
\bottomrule
\end{tabular}
\end{table}

\subsection{TCLab System}

Tables~\ref{tab:glm_tclab_test_loss}--\ref{tab:glm_tclab_time} present the complete GLM coefficient estimates for the TCLab system, covering normalized test loss, train-test gap, and convergence time outcomes respectively.

\begin{table}[htbp]
\centering
\caption{GLM (Gamma) significant coefficients for normalized test loss---TCLab. Reference levels: Adam optimizer, Finetune initialization, 0.5 hr training data.}
\label{tab:glm_tclab_test_loss}
\small
\begin{tabular}{lrrrr}
\toprule
Effect & Coef. & Std.Err. & $z$ & $p$ \\
\midrule
Intercept & 1.554 & 0.131 & 11.89 & $1.32\times10^{-32}$ \\
Retrain & 0.384 & 0.185 & 2.08 & $3.79\times10^{-2}$ \\
12 hr & $-$1.076 & 0.185 & $-$5.82 & $5.82\times10^{-9}$ \\
24 hr & $-$1.205 & 0.185 & $-$6.52 & $7.00\times10^{-11}$ \\
4 hr & $-$0.792 & 0.185 & $-$4.28 & $1.84\times10^{-5}$ \\
SEKF $\times$ 12 hr & 1.081 & 0.261 & 4.13 & $3.57\times10^{-5}$ \\
SEKF $\times$ 24 hr & 0.874 & 0.261 & 3.34 & $8.31\times10^{-4}$ \\
SEKF $\times$ 4 hr & 0.790 & 0.261 & 3.02 & $2.50\times10^{-3}$ \\
SEKF $\times$ Retrain $\times$ 12 hr & $-$1.337 & 0.370 & $-$3.62 & $3.00\times10^{-4}$ \\
SEKF $\times$ Retrain $\times$ 24 hr & $-$1.023 & 0.370 & $-$2.77 & $5.67\times10^{-3}$ \\
L-BFGS $\times$ Retrain $\times$ 4 hr & 0.742 & 0.370 & 2.01 & $4.47\times10^{-2}$ \\
SEKF $\times$ Retrain $\times$ 4 hr & $-$0.795 & 0.370 & $-$2.15 & $3.14\times10^{-2}$ \\
\bottomrule
\end{tabular}
\end{table}

\begin{table}[htbp]
\centering
\caption{GLM (Gamma) significant coefficients for train-test gap---TCLab. Reference levels: Adam optimizer, Finetune initialization, 0.5 hr training data.}
\label{tab:glm_tclab_gap}
\small
\begin{tabular}{lrrrr}
\toprule
Effect & Coef. & Std.Err. & $z$ & $p$ \\
\midrule
Intercept & 1.210 & 0.091 & 13.35 & $1.19\times10^{-40}$ \\
Retrain & 0.288 & 0.128 & 2.25 & $2.45\times10^{-2}$ \\
12 hr & $-$0.657 & 0.128 & $-$5.13 & $2.95\times10^{-7}$ \\
24 hr & $-$0.738 & 0.128 & $-$5.76 & $8.62\times10^{-9}$ \\
4 hr & $-$0.541 & 0.128 & $-$4.22 & $2.42\times10^{-5}$ \\
SEKF $\times$ 12 hr & $-$0.388 & 0.181 & $-$2.14 & $3.24\times10^{-2}$ \\
SEKF $\times$ 4 hr & $-$0.441 & 0.181 & $-$2.43 & $1.50\times10^{-2}$ \\
\bottomrule
\end{tabular}
\end{table}

\begin{table}[htbp]
\centering
\caption{GLM (Gamma) significant coefficients for normalized convergence time---TCLab. Reference levels: Adam optimizer, Finetune initialization, 0.5 hr training data.}
\label{tab:glm_tclab_time}
\small
\begin{tabular}{lrrrr}
\toprule
Effect & Coef. & Std.Err. & $z$ & $p$ \\
\midrule
Intercept & $-$4.168 & 0.449 & $-$9.28 & $1.65\times10^{-20}$ \\
SEKF & $-$1.800 & 0.635 & $-$2.83 & $4.59\times10^{-3}$ \\
SEKF $\times$ Retrain & 2.025 & 0.898 & 2.25 & $2.42\times10^{-2}$ \\
SEKF $\times$ 12 hr & $-$2.197 & 0.898 & $-$2.45 & $1.44\times10^{-2}$ \\
SEKF $\times$ Retrain $\times$ 12 hr & 3.106 & 1.270 & 2.45 & $1.45\times10^{-2}$ \\
\bottomrule
\end{tabular}
\end{table}

\end{document}